# Формализация принципов программирования мозга (Brain Principles Programming)

Витяев Е.Е.[12[0000-0002-5651-6781]], Колонин А.Г.[2[0000-0003-4180-2870]], Андрей Курпатов[3[0000-0003-0431-0543]], Молчанов А.А[3[0000-0003-0197-8871]]

[1] Институт математики им. С.Л.Соболева, пр. Коптюга 4, Новосибирск, Россия
[2] Новосибирский государственный университет, Пирогова 2, Новосибирск, Россия
[3] ПАО Сбербанк, Лаборатория нейронаук и поведения человека, Москва, Россия
`vityaev@math.nsc.ru`

**Абстракт.** В монографии "Сильный искусственный интеллект. На подступах к сверхразуму" содержится обзор общего искусственного интеллекта (AGI). В качестве антропоморфного направления исследований, он включает Brain Principles Programming (BPP) – формализацию универсальных механизмов (принципов) работы мозга с информацией, которые реализуются на всех уровнях организации нервной ткани. В этой монографии содержится формализация этих принципов в терминах теории категорий. Однако этой формализации недостаточно для разработки алгоритмов работы с информацией. В данной работе для описания и моделирования BPP предлагается применять разработанные нами ранее математические модели и алгоритмы, моделирующие когнитивные функции, которые основаны на известных физиологических, психологических и других естественнонаучных теориях. В работе используются математические модели и алгоритмы следующих теорий: Теории Функциональных Систем работы мозга П.К.Анохина, прототипической теории категоризации Eleanor Rosch, теории причинных моделей Bob Rehder и "естественная" классификация. В результате получена формализация BPP и приведены компьютерные эксперименты, демонстрирующие работу алгоритмов.

## 1    Введение

В монографии «Сильный искусственный интеллект. На подступах к сверхразуму» [1] приводится первое кросс-дисциплинарное исследование по общему искусственному интеллекту, где говорится, что «Общий искусственный интеллект – это следующая ступень в развитии ИИ, не обязательно наделенная самосознанием, но, в отличие от современных нейросетей, способная справляться с широким кругом задач в разных условиях». В качестве антропоморфного направления исследований в ней рассматривается Brain

Principles Programming (BPP) – формализация универсальных механизмов (принципов) работы мозга с информацией, сформулированная А.В. Курпатовым. В книге приводится формализация этих принципов в языке теории категорий. Однако, из этой формализации не следуют алгоритмы работы с информацией.

В данной работе сделана попытка применить разработанные ранее математические модели и алгоритмы, моделирующие когнитивные функции и опирающиеся на известные физиологические, психологические и другие естественно-научные теории, для описания и моделирования Brain Principles Programming. Мы будем опираться на следующие теории: Теорию Функциональных Систем работы мозга П.К.Анохина [2-4], теорию интегрированной информации G.Tononi [5], прототипическую теорию категоризации Eleonor Rosh [6-8], теорию причинных моделей Bob Rehter [9-11] и работы по «естественной» классификации [12].

В части I мы приведем математические модели и алгоритмы этих теорий, а затем в части II формализацию BPP в терминах этих моделей и алгоритмом, опираясь на формализацию BPP в терминах теории категорий, приведенную в [1].

Начнем формализацию с некоторых элементарных единиц восприятия внешнего мира. В прототипической теории категоризации и теории «естественных» понятий Eleonor Rosh ими являются «естественные» понятия и прототипы классов, в теории причинных моделей Bob Rehter – причинные модели, в теории интегрированной информацией G.Tononi – концепты, формирующиеся в сознании в виде систем причинных связей с высоко интегрированной информацией.

Эти единицы восприятия описываются в этих подходах как субъективные единицы, однако есть теория, которая описывает объективные свойства объектов, проявляющиеся в этих единицах восприятия – это «естественная» классификация. Далее мы приводим краткое описание этих теорий, начиная с «естественной» классификации и показываем, что все они могут быть формализованы с помощью вероятностных формальных понятий, приводимых далее. Затем приводится метод обнаружения вероятностных формальных понятий.

## Часть I. Базовые теории и формальные модели

## 2    Базовые элементы восприятия, сознания и мира

**2.1. «Естественная» классификация.** Она описывает способ «естественного» формирования понятий об объектах внешнего мира и, как будет показано далее, соответствует исследованиям по формированию «естественных» понятий в когнитивных науках. «Естественная» классификация опирается на объективные свойства внешнего мира и позволяет понять процесс отражения реальности в субъективном опыте.

Первый достаточно подробный философский анализ «естественной» классификации принадлежит Дж. Ст. Миллю [13]. По Дж. Ст. Миллю

«искусственные» классификации отличаются от «естественных» тем, что они могут быть основаны на любом одном или нескольким признаках, так что разные классы различаются только тем, что включают объекты, обладающие различными значениями этих признаков. Но если рассмотреть классы «животных» или «растений», то они отличаются столь большим (потенциально бесконечным) количеством свойств, что их нельзя перечислить. И все эти свойства будут основаны на утверждениях, подтверждающих это различие.

Дж. Ст. Милль дает следующее определение «естественной» классификации: это такая классификация, которая … основывается на таких свойствах, которые служат причинами многих других или по крайней мере составляют их верные признаки. Он определяет также понятие «образа» класса, которое является предтечей «естественных» понятий в когнитивных науках: «наше понятие о классе – тот образ, которым этот класс представлен в нашем уме, – есть понятие о некотором образце, обладающем всеми признаками данного класса … в самой высокой степени».

Рассуждения Дж. Ст. Милля были подтверждены естествоиспытателями. О схожести свойств у «естественных» классов пишет Рутковский Л. [14]: «Чем в большем числе существенных признаков сходны сравниваемые предметы, тем вероятнее их одинаковость и в других отношениях». Смирнов Е.С. [15]: «Таксономическая проблема заключается в "индикации": от бесконечно большого числа признаков нам нужно перейти к ограниченному их количеству, которое заменило бы все остальные признаки».

Из исследований по «естественной» классификации следует, что признаки в «естественных» классах сильно коррелированы. Например, если у нас есть 128 классов и признаки двоичные, то независимыми «индикаторными» признаками среди них могут быть только 7 признаков, потому что $2^7 = 128$, а остальные 121 признаков, в силу схожести свойств по Рутковскому, могут быть предсказаны по этим 7-и признакам, что означает наличие для них 121 закономерности. Так как «индикаторными» признаками могут быть разные 7 признаков, выбранные из 128, и для каждого набора выбранных 7-и признаков есть 121 закономерность, предсказывающая все остальные признаки, то общее число закономерностей может быть не намного меньше, чем $121 \cdot C_{128}^7 = 1143762121920$. В работе [12] приведена формальная модель «естественной» классификации и пример её построения.

**2.2. «Естественные» понятия и прототипическая теория категоризации**. Высокая коррелированность признаков для «естественных» классов была подтверждена в когнитивных исследованиях. В работах Eleanor Rosch [6-8] был сформулирован следующий принцип категоризации «естественных» категорий, подтверждающий высказывания Дж. Ст. Милля и естествоиспытателей: «Perceived World Structure. … perceived world – is not an unstructured total set of equiprobable co-occurring attributes. Rather, the material objects of the world are perceived to possess … ***high correlational structure*** (Курсив ЕЕ). … In short, combinations of what we perceive as the attributes of real objects do not occur uniformly. Some pairs, triples, etc., are quite probable, appearing in combination sometimes with one, sometimes another attribute; others are rare; others logically cannot or empirically do not occur».

Непосредственно воспринимаемые объекты (basic objects) – информационно богатые связки наблюдаемых и функциональных свойств, которые образуют естественную разрывность, создающую категоризацию. Эти связки формируют «прототипы» объектов классов (образ у Дж. Ст. Милля). В дальнейшем теория «естественных» понятий Eleanor Rosch получила название прототипической теории понятий (prototype theory).

**2.3. Теория причинных моделей**. В дальнейших исследованиях было обнаружено, что моделей, основанных на признаках, сходстве и прототипах, недостаточно для описания классов. Необходимо учитывать теоретические, причинные и онтологические знания, относящиеся к объектам классов. Например, люди не только знают, что птицы имеют крылья, могут летать и вить гнезда на деревьях, но также и то, что птицы вьют гнезда на деревьях, потому что могут летать, и летать, потому что имеют крылья.

Исследования показали, что знания людей о категориях не сводится к перечню свойств, а включает богатое множество причинных связей между этими свойствами. Важность свойств категории зависит от их причинных взаимосвязей. В некоторых экспериментах [11] было показано, что свойство важнее, если оно сильнее включено в причинную сеть взаимосвязей признаков.

Учитывая эти исследования, Bob Rehder выдвинул теорию причинных моделей (causal-model theory), в соответствии с которой отношение объекта к категории основывается уже не на множестве признаков и близости по признакам, а на основании сходства порождающего причинного механизма [9]. Для представления причинного знания были использованы Байесовские сети [10]. Однако они не могут моделировать циклические причинные связи, потому что Байесовские сети не поддерживают циклов. Разработанные нами вероятностные формальные понятия, приведенные далее, непосредственно моделируют циклические причинные связи с помощью неподвижных точек предсказаний по причинным связям.

**2.4. Теория интегрированной информации G.Tononi**. Сознание как интегрированная информация. Если «естественная» классификация описывает объекты внешнего мира, а когнитивные науки – восприятие объектов внешнего мира, то теория интегрированной информации сознания G.Tononi анализирует информационные процессы мозга по восприятию объектов внешнего мира.

G.Tononi определяет сознание как первичное понятие, которое обладает следующими феноменологическими свойствами: composition, information, integration, exclusion [5]. Для более точного определения этих свойств G.Tononi вводит понятие интегрированной информации: «это информация, генерируемая системой, которая приходит в определенное состояние после причинно-следственного взаимодействия между ее частями, которая превосходит информацию, генерируемую независимо самими ее частями» [5]. В терминах интегрированной информации феноменологические свойства формулируются следующим образом. В скобках мы приводим интерпретацию этих свойств с точки зрения «естественной» классификации.

- composition – elementary mechanisms (causal interactions) can be combined into higher-order ones («естественные» классы формируются в виде причинных циклов и иерархии «естественных» классов);
- information – only mechanisms that specify 'differences that make a difference' within a system count (только система «резонирующих» причинных связей, формирующая класс, является значимой. См. иллюстрацию на примере ниже);
- integration – only information irreducible to non-interdependent components counts (значима только система «резонирующих» причинных связей, свидетельствующая об избытке информации и восприятии высоко коррелированной структуры «естественного» объекта);
- exclusion – only maxima of integrated information count (только значения признаков, которые максимально взаимосвязаны причинными связями формируют «образ» или «прототип»).

Поскольку у G.Tononi нет внешнего мира и его «естественной» классификации, то приведенные свойства определяются как внутренние свойства системы. Мы рассмотрим эти свойства не как внутренние свойства системы, а как способность системы отражать комплексы причинных связей объектов, а сознание – как способность комплексного иерархического отражения «естественной» классификации внешнего мира.

## 3 Вероятностные формальные понятия и их обнаружение

Нами выдвигается гипотеза о том, что «естественная» классификация, «естественные» понятия и интегрированная информация G.Tononi описываются одним и тем же формализмом. С нашей точки зрения информационные процессы работы мозга и сознание настроилась в процессе эволюции на извлечение высоко коррелированной структуры признаков «естественных» объектов путем формирования «естественных» понятий объектов. Мозг с помощью интегрированной информации настраивается на восприятие «естественных» объектов внешнего мира, отражая их высоко коррелированную структуру. Причинные связи при восприятии «естественных» объектов замыкаются на себя образуя определенный «резонанс», что является системой с высоко интегрированной информацией в смысле G.Tononi. При этом, «резонанс» возникает тогда и только тогда, когда эти причинные связи отражают некоторый целостный «естественный» объект, в котором потенциально бесконечное множество признаков взаимно предсказывают друг друга. Возникающие при этом циклы выводов по причинным связям математически описываются «неподвижными точками» взаимно предсказывающихся свойств, что дает «образ» класса и «прототип» объекта. Поэтому мозг воспринимает «естественный» объект не набором признаков, а как «резонирующую» систему причинных связей. Ниже приведен пример моделирования обнаружения «естественных» классов, «естественных» понятий и интегрированной информации на примере закодированных цифр.

Приведем формализацию циклических причинных связей в виде вероятностных формальных понятий [16-18]. Одновременно будет дано определение Максимально Специфических Вероятностных Причинных Связей (МСВПС), для которых доказано, что логический вывод по ним и, соответственно, вывод предсказаний непротиворечив [19-20]. В теории G.Tononi явно не сказано, каким биологическим субстратом моделируются причинные связи – нейронами, кортикальными колонками или как-то ещё. Мы будем предполагать, что причинные связи обнаруживаются нейронами в соответствии с формальной моделью нейрона, изложенной в [21], которая обнаруживает МСВПС. Кроме того, определенные ниже МСВПС удовлетворяют определению Cartwright [22] вероятностной причинности. «Резонанс» причинных связей в виде неподвижных точек предсказаний по МСВПС причинным связям дан ниже в определении 18, что сразу же приводит к вероятностным формальным понятиям в определении 19. Как будет показано далее, вероятностные формальные понятия в тоже время моделируют понятие контекста.

Приведенная здесь формализация вероятностных формальных понятий следует работам [16-18,23].

**Определение 1.** *Формальный контекст* $K = (G, M, I)$ представляет собой тройку, где $G$ и $M$ – произвольные наборы объектов и атрибутов, и $I \subseteq G \times M$ – бинарное отношение, выражающее принадлежность атрибута объекту.

В формальном контексте операторы производных связывают подмножества объектов и атрибутов контекста.

**Определение 2.** $A \subseteq G, B \subseteq M$, тогда:

$$A^{\uparrow} = \{m \in M \mid \forall g \in A, (g, m) \in I\}$$
$$B^{\downarrow} = \{g \in G \mid \forall m \in B, (g, m) \in I\}$$

**Определение 3.** Пара $(A, B)$ – формальное понятие, если $A^{\uparrow} = B$ и $B^{\downarrow} = A$.

Переопределим контекст в логических терминах. Будем рассматривать только конечные контексты.

**Определение 4.** Для контекста $K = (G, M, I)$ определяем сигнатуру $\Omega_K$ контекста, которая содержит символы предикатов $m(x)$ для каждого $m \in M$, $K \vDash m(x) \Leftrightarrow (x, m) \in I$.

**Определение 5.** Для сигнатуры $\Omega_K$ определим следующий вариант логики первого порядка:

1. $X_K$ – множество переменных;
2. $\mathrm{At}_K$ – множество атомарных формул (атомов) $m(x)$, $m \in \Omega_K$, $x \in X_K$;
3. $L_K$ – набор литералов, включающий атомы $m(t)$ и их отрицания $\neg m(t)$;
4. $\Phi_K$ – набор формул, определяемый индуктивно: литерал – формула, для $\Phi, \Psi \in \Phi_K$ выражения $\Phi \wedge \Psi$, $\Phi \vee \Psi$, $\Phi \rightarrow \Psi$, $\neg \Phi$ – также формулы.

Определим коньюнкцию $\wedge L$ и отрицание $\neg L = \{\neg P \mid P \in L\}$ набора литералов $L \subseteq L_K$.

**Определение 6.** Единичный элемент $\{g\}$, $g \in G$ представленный в сигнатуре $\Omega_K$ образует модель $\mathrm{K}_g$ этого объекта. Истинность формулы $\phi$ на модели $\mathrm{K}_g$ определяется как $g \vDash \phi \Leftrightarrow K_g \vDash \phi$.

**Определение 7.** Определим *вероятностную меру* $\mu$ на множестве G в смысле Колмогорова. Тогда мы можем определить вероятностную меру на множестве формул как:

$\nu : \Phi_K \to [0,1]$, $\nu(\phi) = \mu(\{g \mid g \vDash \phi\})$.

Мы предполагаем, что в контексте нет несущественных объектов, таких что $\mu(\{g\}) = 0$, $g \in G$.

**Определение 8.** Пусть $\{H_1, H_2, \ldots, H_k, C\} \in \mathrm{L}_K$, $C \notin \{H_1, H_2, \ldots H_k\}$, $k \geq 0$.

*Отношение* есть $R = (H_1 \wedge H_2 \wedge \ldots \wedge H_k \to C)$;

*Посылка* $R^{\leftarrow}$ отношения R – это набор литер $\{H_1, H_2, \ldots, H_k\}$;

*Заключение* отношения это $R^{\rightarrow} = C$;

*Длина* отношения это $|R^{\leftarrow}|$;

**Определение 9.** *Вероятность* $\eta$ отношения R – это величина

$$\eta(R) = \nu(R^{\rightarrow} \mid R^{\leftarrow}) = \nu(R^{\leftarrow} \wedge R^{\rightarrow}) \Big/ \nu(R^{\leftarrow}).$$

Если знаменатель $\nu(R^{\leftarrow})$ отношения равен 0, то вероятность не определена.

**Определение 10.** Отношение $\mathrm{R}_1$ является *подотношением* отношения $\mathrm{R}_2$, обозначается как $R_1 \sqsubset R_2$, если $R_1^{\rightarrow} = R_2^{\rightarrow}$, $R_1^{\leftarrow} \subset R_2^{\leftarrow}$.

**Определение 11.** Отношение $\mathrm{R}_1$ *уточняет* отношение $\mathrm{R}_2$, обозначим как $R_2 < R_1$, если $R_2 \sqsubset R_1$ и $\eta(R_1) > \eta(R_2)$.

**Определение 12.** Отношение R является *вероятностной причинной связью*, если для каждого $\tilde{R}$ выполнено $(\tilde{R} \sqsubset R) \Rightarrow (\tilde{R} < R)$.

Определение вероятностной причинности, данное Cartwright [22] относительно некоторого бэкграунда, может быть сформулировано в приведенных терминах следующим образом. Если посылкой $R^{\leftarrow}$ отношения R является набором литералов $\{H_1, H_2, \ldots, H_k\}$ и мы рассматриваем этот набор как бэкграунд, то каждый литерал посылки является вероятностной причиной заключения $R^{\rightarrow}$ отношения R относительно этого бэкграунда, то есть

$\nu(R^{\rightarrow} / R^{\leftarrow}) > \nu(R^{\rightarrow} / (R^{\leftarrow} \setminus H))$ для каждого $\mathrm{H} \in \{H_1, H_2, \ldots, H_k\}$.

Легко видеть, что это определение следует из определения 12.

**Определение 13.** *Сильнейшей вероятностной причинной связью* будет называться отношение R, для которого не существует такой вероятностной причинной связи $\tilde{R}$, что $(\tilde{R} > R)$.

**Определение 14.** *Семантический Вероятностный Вывод* (СВВ) предсказаний некоторого литерала C есть последовательность вероятностных причинных связей $R_0 < R_1 < R_2 ... < R_m$, $R_0^{\rightarrow} = R_1^{\rightarrow} = R_2^{\rightarrow} ... = R_m^{\rightarrow} = C$, $R_0^{\leftarrow} = \varnothing$, $R_m$ – сильнейшая вероятностная причинная связь.

**Определение 15.** *Дерево семантического вероятностного вывода* Tree(C) некоторого литерала C – это совокупность всех СВВ, предсказаний литерала C.

**Определение 16.** *Максимально специфичное причинное отношение* для предсказания некоторого C – это сильнейшее вероятностное причинное отношение дерева Tree(C), имеющее максимальную условную вероятность.

Обозначим через MSCR множество всех максимально специфичных причинных отношений. Под *системой причинных отношений* будем понимать любое подмножество $\mathcal{R} \subseteq$ MSCR.

**Определение 17.** Определим *оператор предсказания* для системы $\mathcal{R}$ как:

$$\Pi_{\mathcal{R}}(L) = L \cup \{C \mid \exists R \in \mathcal{R} : R^{\leftarrow} \subseteq L, R^{\rightarrow} = C\}.$$

**Определение 18.** *Замыканием* набора литералов L назовем наименьшую неподвижную точку оператора предсказания, содержащую L:

$$\Pi_{\mathcal{R}}^{\infty}(L) = \bigcup_{k \in \mathbb{N}} \Pi_{\mathcal{R}}^{k}(L).$$

Набор литералов L *непротиворечив*, если он не содержит одновременно атом C и его отрицание $\neg C$. Набор литералов L *совместен*, если $\nu(\wedge L) \neq 0$.

**Теорема 1.** [18-19]. Если L – совместно, то $\Pi_{\mathcal{R}}(L)$ совместно и непротиворечиво для любой системы $\mathcal{R}$.

**Определение 19.** *Вероятностное формальное понятие* на контексте K – это пара (A, B), удовлетворяющая следующим условиям:

$$\Pi_{\mathcal{R}}^{\infty}(B) = B, A = \bigcup_{\Pi_{\mathcal{R}}^{\infty}(C) = B} C^{\downarrow}.$$

Определение множества A основано на следующей теореме, связывающей вероятностные и стандартные формальные понятия на контексте K.

**Теорема 2.** [18-19]. Пусть $K = (G, M, I)$ – формальный контекст, тогда:

5. Если (A,B) – формальное понятие на K, то существует вероятностное формальное понятие (S,T) на K такое, что $A \subseteq S$, $B \subseteq T$.

6. Если (S,T) – вероятностное формальное понятие на K, то существует семейство $\mathcal{C}$ формальных понятий на K, такое что

$$\forall (A, B) \in \mathcal{C} \ (\Pi_{\mathcal{R}}^{\infty}(B) = T), S = \bigcup_{(A,B) \in \mathcal{C}} A.$$

# 4 Алгоритм статистической аппроксимации вероятностных формальных понятий.

В практических задачах мы не можем предполагать, что вероятностная мера нам известна. Поэтому, нам необходимо использовать некоторый статистический критерий для определения вероятностных неравенств в семантическом вероятностном выводе и обнаружении МСВПС [12,24]. Для этого мы используем точный критерий независимости Фишера с уровнем значимости $\alpha$.

Результирующий набор $\mathcal{R}_\alpha$ вероятностных максимально специфических причинно-следственных связей, полученный с уровнем значимости $\alpha$, может вызывать противоречия в неподвижных точках вероятностных формальных понятий. Следовательно, для аппроксимации оператора $\Pi_\mathcal{R}(L)$ необходимо ввести дополнительный критерий согласованности максимально специфических причинных связей $\mathcal{R}_\alpha$ на множестве L.

**Определение 20.** Причинное отношение $R \in \mathcal{R}_\alpha$ *подтверждается* на множестве литералов $L$, если $R^\leftarrow \subset L$ и $R^\rightarrow \in L$. Тогда $R \in \mathrm{Sat}(L) \subseteq \mathcal{R}_\alpha$.

**Определение 21.** Причинное отношение $R \in \mathcal{R}_\alpha$ *опровергается* на множестве литералов $L$, если $R^\leftarrow \subset L$ и $R^\rightarrow \in \neg L$. Тогда $R \in \mathrm{Fal}(L) \subseteq \mathcal{R}_\alpha$.

Теперь мы можем определить критерий максимальной согласованности предсказаний по максимально специфическим причинным связям $\mathcal{R}_\alpha$ на некотором множестве литералов L.

**Определение 22.** *Критерием максимальной согласованности предсказаний* по максимально специфическим причинным связям $\mathcal{R}_\alpha$ на множестве литералов L является значение:

$$\mathrm{Int}(L) = \sum_{R \in \mathrm{Sat}(L)} \gamma(R) - \sum_{R \in \mathrm{Fal}(L)} \gamma(R).$$

Выбор оценки причинной связи $\gamma$ может зависеть от специфики задачи. В наших экспериментах мы руководствовались соображениями Шеннона:

$$\gamma(R) = -\log(1 + \epsilon - \eta(R)),\ \epsilon > 0, \epsilon \ll 1.$$

Теперь мы можем аппроксимировать оператор $\Pi_\mathcal{R}(L)$, используя критерий согласованности предсказаний.

**Определение 23.** Определим *оператор максимальной согласованности предсказаний* $\Upsilon(L)$ для множества $\mathcal{R}_\alpha$ максимально специфических причинных связей, который аппроксимирует оператор $\Pi_\mathcal{R}(L)$. Он изменяет набор литер $L$ на один элемент так, чтобы строго увеличить критерий $\mathrm{Int}(L)$:

1. Для всех $G \in L_\kappa \setminus L$ вычислить максимальное увеличение критерия от добавления G к $L$: $\Delta^+ = \mathrm{Int}(L \cup \{G\}) - \mathrm{Int}(L)$ при условии, что в $\mathrm{Sat}(L)$ есть закономерность $R \in \mathrm{Sat}(L)$ такая что $R^\leftarrow \subset L$ и $R^\rightarrow = G$.

2. Для всех $G \in L$ вычислить максимальное увеличение критерия от удаления G из L: $\Delta^- = \mathrm{Int}(L \setminus \{G\}) - \mathrm{Int}(L)$;

3. Оператор $\Upsilon(L)$ добавляет литерал G к L, если $\Delta^+ > 0$ и $\Delta^+ > \Delta^-$;

4. Оператор $\Upsilon(L)$ удаляет литерал G из L, если $\Delta^- > 0$ и $\Delta^- > \Delta^+$.

5. Если $\Delta^- = \Delta^+$ и $\Delta^- > 0$, оператор $\Upsilon(L)$ удаляет литерал G;

6. Если $\Delta^+ \leq 0$ и $\Delta^- \leq 0$, оператор $\Upsilon(L)$ возвращает L и, следовательно, мы получили неподвижную точку оператора максимальной согласованности предсказаний.

**Определение 24.** Под *статистической аппроксимацией вероятностных формальных понятий* контекста K для максимально специфических причинных связей $\mathcal{R}_\alpha$ мы понимаем набор всех неподвижных точек $\Upsilon^\infty(L)$, которые могут быть получены в результате многократного применения оператора $\Upsilon(L)$ к некоторому набору литералов L, представляющему некоторый объект $L = \{g\}^\uparrow$.

Докажем, что в предельном случае, когда множество закономерностей $\mathcal{R}_\alpha$ совпадает с системой причинных отношений $\mathcal{R} \subseteq \text{MSCR}$, неподвижная точка оператора предсказания $\Pi_\mathcal{R}^\infty(B)$ и оператора максимальной согласованности предсказаний $\Upsilon(L)$ совпадают. Поэтому статистическая аппроксимация вероятностных формальных понятий является прямым обобщением исходных вероятностных формальных понятий на случай работы с зашумленными данными.

**Теорема 3.** Пусть $\mathcal{R}_\alpha = \mathcal{R} \subseteq \text{MSCR}$. Тогда для любого совместного набора литер L $\Upsilon^\infty(L) = \Pi_\mathcal{R}^\infty(L)$.

Доказательство: В силу теоремы 1 для $\mathcal{R}_\alpha = \mathcal{R}$ и L совместного у нас всегда будет $\text{Fal}(L) = \varnothing$ на любом шаге применения оператора $\Upsilon(L)$. Тогда, поскольку $\gamma(R) = -\log(1 + \epsilon - \eta(R)) = -\log(\epsilon) > 0$, то $\text{Int}(L) = \sum_{R \in \text{Sat}(L)} \gamma(R) > 0$, при $\text{Sat}(L) \neq \varnothing$.

Тогда всегда будет выполняться неравенство $\Delta^- = \text{Int}(L \setminus \{G\}) - \text{Int}(L) \leq 0$ поскольку $\text{Sat}(L \setminus G) \subseteq \text{Sat}(L)$. Поэтому в соответствии с определением 23 оператор $\Upsilon(L)$ не будет удалять литеры из $L$.

С другой стороны, оператор $\Upsilon(L)$ будет добавлять новые литеры $G$ в $L$ при условии, что $\Delta^+ = \text{Int}(L \cup \{G\}) - \text{Int}(L) > 0$. Это значит, что $\text{Sat}(L) \subset \text{Sat}(L \cup \{G\})$ и существует закономерность $R \in \text{Sat}(L \cup \{G\})$ такая что $R^\leftarrow \subseteq L$ и $R^\rightarrow = G$. Это означает, что оператор $\Upsilon(L)$ всегда будет добавлять к $L$ одну из литер $G$ для которой существует закономерность $R^\leftarrow \subseteq L$ и $R^\rightarrow = G$.

Таким образом, в нашем случае оператор $\Upsilon(L)$ можно записать так:

$$\Upsilon(L) = L \cup \frac{\arg\max \eta(G)}{\{G \mid \exists R \in \mathcal{R} : R^\leftarrow \subseteq L, R^\rightarrow = G,\}}.$$

Напомним, что оператор предсказания имеет аналогичный вид:
$$\Pi_\mathcal{R}(L) = L \cup \{C \mid \exists R \in \mathcal{R} : R^\leftarrow \subseteq L, R^\rightarrow = C\}.$$

Отличие операторов состоит только в последовательности добавления литер. Однако они всегда добавляют к множеству $L$ те и только те литералы $G$, для которой существует отношение $R \in \mathcal{R}$ такое что $R^\leftarrow \subset L$ и $R^\rightarrow = G$. Поскольку порядок добавления литер не влияет на возможность включения других литер, то получаемые в результате неподвижные точки $\Upsilon^{\infty}(L)$ и $\Pi_{\mathcal{R}}^{\infty}(L)$ совпадают.

## 5    Обнаружение «естественных» понятий и контекстов

Приведем пример работы алгоритма статистической аппроксимации вероятностных формальных понятий для некоторого контекста $K = (G, M, I)$, где G – это множество закодированных цифр как показано на рис. 1, M – это множество признаков цифр (см. рис. 1а) и I отношение, связывающее признаки и цифры. Для эксперимента было взято множество из 360 перетасованных цифр (12 цифр рис. 1 продублированных в 30-ти экземплярах без указания, где какая цифра). На этом множестве было обнаружено множество $\mathcal{R}_{\alpha}$ из 55089 вероятностных максимально специфических причинно-следственных связей, полученных с уровнем значимости $\alpha = 0.01$. Пусть L – это множество литералов, определенных для всех значений всех признаков. По причинно-следственным связям $\mathcal{R}_{\alpha}$ оператором $\Upsilon(L)$ было обнаружено 12 статистических аппроксимаций вероятностных формальных понятий точно соответствующих 12-и цифрам.

Пример неподвижной точки для цифры 6 приведен на рис. 2. Рассмотрим, что представляет собой эта неподвижная точка. Занумеруем признаки цифр, как указано на рис. 1. Первая закономерность цифры 6, представленная в первом прямоугольнике после фигурной скобки означает, что, если в квадрате 13 стоит признак 6 (обозначим 13-6), то в квадрате 3 должен стоять признак 2 (обозначим как (3-2)). Предсказываемый признак обозначается точечной линией. Запишем это отношение как (13-6 $\Rightarrow$ 3-2). Нетрудно проверить, что это отношение выполнена на всех цифрах. Второе отношение означает, что из признака (9-5) и отрицания значения 5 первого признака $\neg$(1-5) (первый признак не должен быть равен 5) следует признак (4-7). Отрицание обозначается на рисунке пунктирной линией, как показано в нижней части рис. 2. Получим

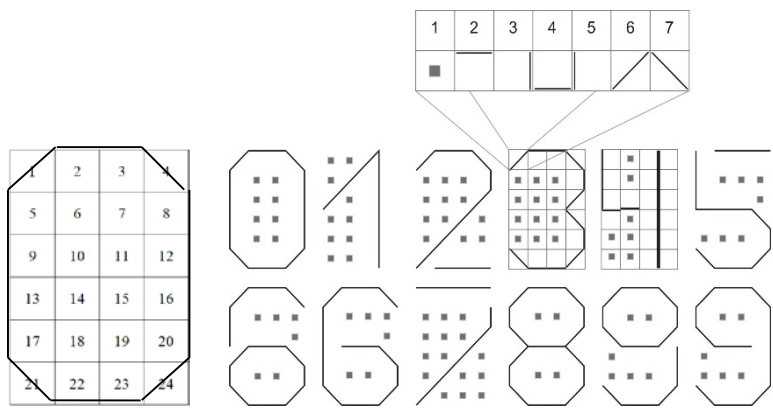

**Рис. 1.** Кодировка цифр

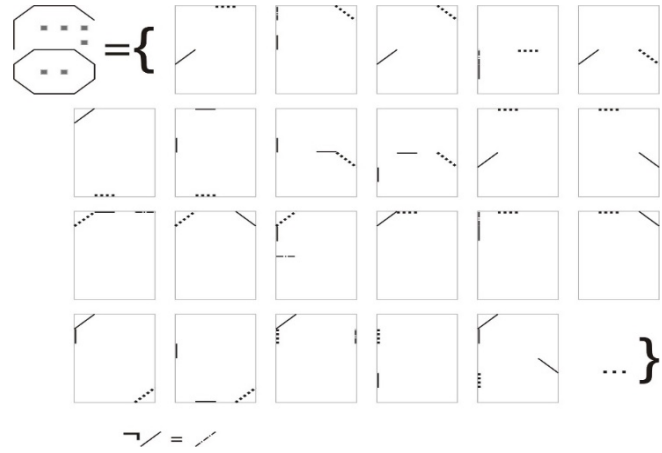

**Рис. 2.** Неподвижная точка цифры 6.

отношение
(9-5&¬(1-5) ⇒ 4-7). Последующие 3 отношения в первой строке цифры 6 будут соответственно (13-6 ⇒ 4-7), (17-5&¬(13-5) ⇒ 4-7), (13-6 ⇒ 16-7).

На рис. 2 видно, что отношения и признаки цифры 6 образуют неподвижную точку – взаимно предсказывают друг друга. Заметим, что отношения, используемые в неподвижной точке, выполнены на всех цифрах, а сама неподвижная точка выделяет только одну цифру. Это иллюстрирует феноменологическое свойство 2 G.Tononi 'differences that make a difference', в котором система причинных связей воспринимает «осознает» целостный объект. Поэтому цифры выделяются не закономерностями самими по себе, а их системной взаимосвязью. Неподвижная точка формирует «прототип» по Eleanor Rosch или «образ» по Дж. Ст. Миллю. Программа не знает заранее, какие сочетания признаков максимально коррелируют между собой.

Важно отметить, что причинные связи в неподвижной точке предсказывают не только наличие некоторого другого признака в этой неподвижной точке, но и невозможность наличия какого-то другого признака в этой неподвижной точке. Таким образом, неподвижная точка характеризуется не только наличием признаков в соответствующих квадратах, но и необходимостью отсутствия признаков в каких-то других квадратах, то есть неподвижная точка моделирует процесс вытормаживания признаков и соответствующих прототипов других классов.

**Формирование контекста на примере анализа социальных сетей**. Покажем, каким образом алгоритм статистической аппроксимации применим для обнаружения контекста. Исследуем данные о 2784 респондентов (пользователей) социальной сети. У нас есть данные о 36 характеристиках пользователей. На рисунке 3 изображены исходные данные.

В результате работы алгоритма статистической аппроксимации была получена 21 неподвижная точка оператора $\Upsilon(L)$, которые в данном случае

| ID | sex | friend* | educat* | relation | is_relativ es* | album* | videos* | audios* | notes* | photo* | group* | user_vid eos* | followers* | use_polit ical* | use_rel* | post_coun* | post_like_su m | use_attitud* |
|---|---|---|---|---|---|---|---|---|---|---|---|---|---|---|---|---|---|---|
| Скрыт полноc Женский | 1 | 0 Не указанo/С | 0 | 3 | 37 | 87 | 0 | 15 | 0 | 0 | | | 491 указано t указано | 407 | 14 Не указано | | | |
| Указана полноc Мужской | 84 | 0 Не указанo/С | 0 | 3 | 15 | 32 | 0 | 22 | 0 | 3 | | | 14 указано t указано | 53 | 72 Не указано | | | |
| Указана полноc Мужской | 47 | 0 Не указанo/С | 0 | 5 | 20 | 93 | 0 | 122 | 7 | 0 | | | 10 указано t указано | 44 | 47 Не указано | | | |
| Скрыт под рожд Мужской | 0 | 1 Не указанo/С | 0 | 0 | 23 | 1 | 0 | 0 | 0 | 0 | | | 13 указано t указано | | 0 Не указано | | | |
| Указана полноc Мужской | 153 | 1 женат/замуж | 1 | 0 | 1 | 2 | 0 | 38 | 32 | 0 | | | 1 указано t указано | 32 | 81 Не указано | | | |
| Скрыт под рожд Женский | 16 | 0 женат/замуж | 1 | 0 | 6 | 0 | 0 | 24 | 1 | 0 | | | 4 указано t указано | 3 | 3 Не указано | | | |
| Скрыт под рожд Мужской | 1 | 0 Не указанo/С | 0 | 0 | 61 | 5 | 0 | 5 | 0 | 0 | | | 23 указано t указано | 4 | 23 Не указано | | | |
| Скрыт под рожд Мужской | 67 | 0 Не указанo/С | 0 | 0 | 701 | 0 | 0 | 208 | 0 | 0 | | | 140 указано t указано | 89 | 105 Не указано | | | |
| Указана полноc Мужской | 5 | 1 женат/замуж | 1 | 0 | 0 | 228 | 0 | 0 | 0 | 0 | | | 3 указано t указано | 4 | 0 Не указано | | | |
| Скрыт под рожд Женский | 19 | 1 Не указанo/С | 0 | 0 | 0 | 1 | 3 | 0 | 0 | 0 | | | 3 указано t указано | | 0 Не указано | | | |
| Скрыт под рожд Женский | 38 | 1 Не указанo/С | 0 | 2 | 9 | 164 | 0 | 28 | 6 | 0 | | | 9 указано | 6 | 18 Указано | | | |
| Указана полноc Мужской | 36 | 1 Не указанo/С | 0 | 0 | 2 | 0 | 0 | 0 | 0 | 0 | | | 3 указано t указано | 1 | 0 Не указано | | | |
| Указана полноc Мужской | 53 | 1 не женат/нe | 0 | 2 | 130 | 0 | 0 | 36 | 32 | 0 | | | 16 Указано | 41 | 51 Не указано | | | |
| Указана полноc Женский | 209 | 1 женат/замуж | 1 | 5 | 1142 | 0 | 0 | 54 | 113 | 6 | | | 86 Указано | 155 | 165 Не указано | | | |
| Скрыт под рожд Мужской | 4 | 0 не женат/нe | 0 | 0 | 0 | 0 | 0 | 11 | 1 | 0 | | | 0 указано t указано | 11 | 0 Не указано | | | |
| Указана полноc Мужской | 7 | 0 не женат/нe | 0 | 0 | 0 | 0 | 0 | 0 | 0 | 0 | | | 2 указано t указано | 4 | 0 Не указано | | | |
| Скрыт полноc Мужской | 35 | 0 Не указанo/С | 0 | 0 | 0 | 2 | 0 | 2 | 0 | 0 | | | 3 указано t указано | | 0 Не указано | | | |
| Указана полноc Мужской | 181 | 1 женат/замуж | 1 | 0 | 69 | 13 | 0 | 6 | 0 | 0 | | | 23 t указано t указано | 821 | 100 Указано | | | |
| Указана полноc Мужской | 149 | 0 Не указанo/С | 0 | 0 | 53 | 258 | 2 | 37 | 0 | 114 | | | 8 t указано t указано | 61 | 204 Не указано | | | |
| Указана полноc Мужской | 82 | 0 Не указанo/С | 0 | 0 | 0 | 0 | 0 | 7 | 8 | 0 | | | 18 t указано t указано | 7 | 10 Не указано | | | |
| Указана полноc Мужской | 542 | 0 Не указанo/С | 0 | 6 | 19 | 0 | 3 | 254 | 0 | 0 | | | 640 t указано t указано | 184 | 225 Не указано | | | |
| Указана полноc Мужской | 26 | 0 Не указанo/С | 0 | 3 | 182 | 0 | 4 | 0 | 0 | 0 | | | 12 t указано t указано | 15 | 0 Не указано | | | |
| Скрыт полноc Мужской | 31 | 1 Не указанo/С | 1 | 2 | 0 | 0 | 0 | 24 | 0 | 0 | | | 10 t указано t указано | 4 | 0 Не указано | | | |
| Скрыт под рожд Мужской | 49 | 0 Не указанo/С | 0 | 0 | 0 | 0 | 0 | 0 | 0 | 0 | | | 10 t указано t указано | 1 | 0 Не указано | | | |
| Скрыт под рожд Женский | 13 | 1 Не указанo/С | 1 | 1 | 1 | 0 | 0 | 28 | 0 | 0 | | | 11 Указано Указано | 5 | 0 Не указано | | | |

Рис. 3. Характеристиках пользователей.

представляют собой контексты или типы пользователей сети. Наиболее характерными из них являются 8-й контекст и 11-ый. 8-й контекст можно охарактеризовать как замужнюю женщину, проживающую в своем родном городе, для которой главное в жизни семья и дети, ценит в людях доброту и честность, в профиле указаны родственники, отношение к алкоголю и курению негативное, небольшое число фото, видео, аудио. 11-й контекст можно охарактеризовать как неженатого мужчину, возможно подростка, который также ценит в людях доброту и честность, но главное для него в жизни – это саморазвитие, компромиссно относится к курению и алкоголю.

## Часть II. Формализация принципов программирования мозга (Brain Principles Programming)

### 6 «Интеллектуальный объект» и «интеллектуальная функция»

В основе формализации «Brain Principles Programming, (BPP)», изложенной в [1] и осуществленной в теории категорий лежат понятия «интеллектуальный объект» и «интеллектуальная функция».

Приведем сначала неформальные определения «интеллектуального объекта» и «интеллектуальной функции» из [1,25]:

- «интеллектуальный объект», под которым мы понимаем любую единичную целостность, выделяемую нами в этом пространстве – например, когда мы видим стол, сигналы от зрительного нерва обрабатываются мозгом, и сочетание отдельных линий опознается как стол;

- «интеллектуальная функция», которая описывает все возможные операции в рассматриваемой системе – это все, что психика может сделать с интеллектуальным объектом. Когда мы опознаем стол как объект, мы можем оценить его размер или придумать, как его использовать;

• «сущность» – специфическое значение объекта для психики. То есть, знание о том, для чего можно использовать стол.

Основная идея формализации этих понятий в рамках теории категорий заключается в утверждении: «некий наблюдатель – субъект опыта, которому является Мир – присваивает значения вещам исключительно через призму взаимодействия с ними» или иначе «вещи, которыми мне является Мир, существуют для меня и лишь в отношении со мной» [1].

Формально «интеллектуальный объект» определяется как отображение, где:
• некоторый набор данных (А);
• наблюдатель, как отражение мира после взаимодействия с ним (Ω);
• отношение мира с наблюдателем, являющееся функцией внутреннего состояния/ожидания (f) – интеллектуальная функция $A \xrightarrow{f} \Omega$ [1].

Здесь А – набор данных, характеризующий любую единичную целостность; Ω – различительная способность: «отношение интеллектуального объекта со «мной» еще не означает какой-либо осознанности, представленности данного интеллектуального объекта в сознании – достаточно того, чтобы нечто было хоть как-то воспринято и распознано в степени, достаточной для того, чтобы это «нечто» в будущем было так или иначе учтено, принято в расчет... поскольку основной задачей мышления, как уже отмечалось выше, является предсказание, или производство конкурентного будущего, то каким нам в итоге представится воспринимаемый объект, будет зависеть от нашей настроенности, или, как сказали бы феноменологи, от нашей интенциональности» [26].

Более формально «К объекту Ω, моделирующему различительную способность, таким образом, будут предъявлены некоторые требования: это должно быть, во-первых, частично-упорядоченное множество, элементам которого соответствуют «более» или «менее» высокие значения. То есть на элементах данного множества должна иметься структура порядка. Иначе говоря, мы будем использовать Ω как некую экзистенциальную меру или, попросту говоря, линейку, которой мы будем измерять различия» [26].

Далее определяется функция ожидания $Exp_A : A \times A \to \Omega$ (от англ. expectation) и даются следующие пояснения этой функции: «на всех уровнях восприятия ... мы по сути имеем дело с ситуацией, с некоторым ожидаемым положением дел ... *Наша психика непреодолимо тяготеет к тому, чтобы сложить весь набор раздражителей в некую понятную, ясную и как бы непротиворечивую картину реальности* (курсив 1 – Е.Е.) ... Эти представления о реальности, в свою очередь, являются специфическим фильтром-интерпретатором – *всякие новые раздражители, оказываясь, образно говоря, в поле тяготения соответствующей системы представлений, неизбежно как бы изменяют свою траекторию – одни отталкиваются (игнорируются), другие, комплементарные, напротив, притягиваются, третьи – видоизменяются (интерпретируются) в угоду господствующим установкам* (курсив 2 – Е.Е.) ... В результате в отношении любого элемента x, входящего в состав интеллектуального объекта А, осмысленно говорить, насколько он, во-первых, отличен от самого себя в смысле того, что мы ожидаем увидеть на его месте, и, во-вторых, насколько он уместен в ситуации вообще, т. е. насколько он близок остальным элементам, различенным в ситуации».

Более формально [26], функция ожидания $Exp_A : A \times A \to \Omega$ сопоставляет каждой паре элементов $x, y \in A$ меру их согласованности (когерентности) на нашей экзистенциальной частично-упорядоченной шкале $\Omega$. При этом мера согласованности объекта $x \in A$ с самим собой $Exp_A(x, x)$ может пониматься как мера близости x к своей сущности ... и обозначаться как $Ess_A(x)$ (от англ. essence). Если считать прототип объектов класса как «инвариант» объектов класса, то неподвижная точка оператора $\Upsilon^\infty(X(y))$, полученная на множестве свойств, заданных предикатами $X(y) = \{P_1 \& ... \& P_m\}$ для некоторого элемента $y \in A$, будет отличаться от признаков $X(y)$ самого элемента в точности как мера согласованности объекта с самим собой. Поэтому оператор $\Upsilon^\infty(X(y))$ дает определенную меру близости объекта к своей сущности (инварианту) «то есть мозг учится неким шаблонам восприятия — формирует в себе некие идеальные (инвариантные данной «сущности») модели, которые впоследствии помогают ему быстро объединять разрозненные данные, чтобы идентифицировать те или иные объекты, как бы вкладывая их в соответствующий инвариант». [25].

Функция ожидания позволяет полнее определить интеллектуальный объект [26]. «Под интеллектуальным объектом мы будем понимать ... объект $\mathcal{A} := (A, Exp_A)$, включающий в себя множество данных A и функцию ожидания $Exp_A : A \times A \to \Omega$, существенным образом зависящую от субъекта опыта $\Omega$ и его внутреннего состояния».

Строение интеллектуального объекта, описанное курсивом 1 выше, фактически означает, что интеллектуальный объект представляет собой контекст, представленный вероятностным формальным понятием, в котором оператор $\Upsilon^\infty(A)$, имеющим тот же смысл – минимизации противоречий в наборе раздражителей – по входному множеству раздражителей A генерирует максимально непротиворечивую картину реальности, дополняя ее всей, соответствующей ситуации информацией. При этом (см. курсив 2) новые раздражители либо меняют свою траекторию, либо отталкиваются, либо притягиваются. Все эти эффекты, которые учитываются в функции ожидания $Exp_A : A \times A \to \Omega$, моделируются взаимодействием вероятностных формальных понятий элементов множества А. Изменение траектории – это перевод признаков ближе к прототипу, отталкивание – это вытормаживание признаков, о котором говорилось выше и притяжение – это взаимная поддержка в неподвижной точке.

**Интеллектуальная функция**. Работа интеллектуальной функции состоит не только в том, чтобы воссоздать интеллектуальные объекты, связанные с элементами данных А и самим множеством А, но и соединить эти интеллектуальны объекты со всеми другими интеллектуальными объектами, имеющимися в психике и имеющими отношение к данной ситуации, например, к потребности, имеющейся в данной ситуации или некоторой задаче (цели) ««мир интеллектуальной функции» — это все возможные «интеллектуальные

объекты», которые могут оказаться в пространстве психического (по существу, речь идет о культурно-историческом содержании, как его понимал Л. С. Выготский). Причем они воспроизводятся конкретной психикой через отношение — интеллектуальную функцию — с другими, уже существующими в ней интеллектуальными объектами» [25].

Итогом работы интеллектуальной функции является создание «тяжелого интеллектуального объекта» путем «укрупнения имеющихся у нас знаний, которые мы полагаем относящимися к некоторой занимающей нас проблеме» [25]. Таким образом, интеллектуальный объект $\mathcal{A} := (A, Exp_A)$ «как бы возводится в степень тех знаний (интеллектуальных объектов), которыми мы обладаем, и обретает для нас соответствующее значение» [26].

Возведение некоторого интеллектуального объекта $\mathcal{A}$ «в степень» знаний формально представляется как отношение: «Если отношение мыслить, как определенного вида направленную связь, то кажется вполне естественным обозначать интеллектуальные объекты буквами $\mathcal{A}, \mathcal{B}, \mathcal{C} \ldots$, а отношения между ними стрелками … $r : \mathcal{A} \to \mathcal{B}$ » [26]. Кроме того, это отношение «должно уважать те различия и отождествления, которые были положены функцией ожидания $Exp_A$» и удовлетворять следующим условиям:

$$\forall a, b \in A (Ess_B(r(a)) \le Ess_A(a)), \ Exp_A(a,b) \le Exp_B(r(a), r(b)).$$

Все стрелки отношения r, показывающие путь обогащения некоторого интеллектуального объекта, образуют «конус». Пределом диаграммы обогащений является конус, порожденный «тяжелым» объектом, содержащимся во всех других конусах.

Оператор $\Upsilon^\infty(A)$ автоматически формирует контекст, порожденный A и вероятностными формальными понятиями ее элементов, поскольку по всем причинным связям, связывающим элементы A с другими знаниями, имеющимися в психике, другие ожидаемые элементы психики автоматически будут включены в контекст по этим причинным связям, если конечно они не будут сильно противоречить имеющейся информации, что проверяется этим оператором. Если, при этом, активируются некоторые высокоуровневые понятия, например, работа, учеба, то не будет извлекаться вся связанная с ними информация, кроме самой общей, а будет извлекаться информация, привязанная к контексту и свойствам ситуации, имеющимся в A. Пределом работы оператора $\Upsilon^\infty(A)$ является в этом случае контекст, соответствующий «тяжелому интеллектуальному объекту».

## 7 Теория Функциональные Систем

Понятия «интеллектуальный объект» и «интеллектуальная функция» в рамках Brain Principles Programming должны описывать когнитивные функции и прежде всего мышление. Покажем на примере ведущей в России физиологической теории целенаправленной деятельности – Теории Функциональных Систем (ТФС), как такая теория может быть описана в терминах «интеллектуальных объектов» и «интеллектуальных функций».

Само по себе мышление целенаправленных действий не предполагает. Мы можем планировать достижений каких-то целей лежа на диване. Поэтому разобьем целенаправленное поведение по удовлетворению некоторой потребности на два этапа – этап планирования действий и принятия решения, который осуществляется еще до всяких действий, как формирование контекста, включающего «образ потребного будущего», и этап осуществления целенаправленного поведения в соответствии с принятым решением вместе с контролем достижения промежуточных и конечного результатов в соответствии с акцептором результатов действий [4].

Когда возникает некоторая потребность, а, как правило, всегда доминирует некоторая потребность, если учесть, что спектр потребностей достаточно широк, то, во-первых, во множестве A должен быть элемент и соответствующий интеллектуальный объект, сформированный мотивационным возбуждением, которое активирует процесс поиска решения по удовлетворению потребности, а, во-вторых, частично-упорядоченное множество $\Omega$, моделирующее нашу различительную способность по ожиданию функцией $Exp_A : A \times A \to \Omega$ удовлетворения нашей потребности, будет оценивать элементы A и их взаимодействие с точки зрения удовлетворения потребности и соответствующим образом влиять на конструирование интеллектуального объекта «образа потребного будущего». Поэтому первый этап можно рассматривать как контекст функциональной системы по удовлетворению потребности, сформированный мотивационным возбуждением и «образом потребного будущего».

В теории функциональных систем достаточно подробно описан процесс формирования «образа потребного будущего». Полная и подробная формализация функциональных систем приведена в работах [27-28].

Любая функциональная система имеет следующую архитектуру, первый этап которой мы проинтерпретируем как формирование контекста.

**Первый этап** включает:

**Афферентный синтез**. Включающий в себя синтез мотивационного возбуждения, памяти, обстановочной и пусковой афферентации:

- **Мотивационное возбуждение**. Постановка цели в целенаправленном поведении осуществляется возникшей потребностью, которая трансформируется в мотивационное возбуждение – возбуждение «центральных мозговых структур», инициируемое возникшей потребностью. Мотивационное возбуждение формирует базовый интеллектуальный объект $\mathcal{M} := (M, Exp_M)$, задающий различительную способность $\Omega$, которая будет определять, что нужно, а что не нужно для удовлетворения потребности. Элементами этого интеллектуального объекта являются возбуждения соответствующих мозговых структур.

- **Память**. Мотивационное возбуждение «извлекает из памяти» все последовательности действий, которые ранее приводили к достижению цели. Таким образом, интеллектуальный объект $\mathcal{M} := (M, Exp_M)$ «возводится в степень» – обогащается опытом тех случаев, которые ранее приводили к удовлетворению данной потребности. В результате получаем множество $\{\mathcal{C} := (C, Exp_C)\}$ обогащенных интеллектуальных объектов, соответствующих каждому случаю.

- **Обстановочная афферентация**. Мотивационное возбуждение с учетом текущей обстановки извлекает из памяти только тот опыт по достижению цели, который возможен в данной обстановке. Поэтому выбираются только интеллектуальные объекты тех способов $\mathcal{C} := (C, Exp_C)$ достижения цели, которые возможны в данной обстановке.
- **Пусковая афферентация**. По смыслу эта афферентация также является обстановочной афферентацией, только она связанна со временем и местом достижения результата.

**Принятие решений**. На стадии афферентного синтеза мотивационным возбуждением может быть извлечено из памяти несколько способов достижения цели и сформировано соответствующее множество $\{\mathcal{C} := (C, Exp_C)\}$ интеллектуальных объектов. В соответствии с формализацией [27-28] эти способы включают в себя правила вида:

$$P_1 \& \dots \& P_n \& PG_1 \& \dots \& PG_m \& A_1 \& \dots \& A_k \Rightarrow PG_0,$$

где $P_1 \& \dots \& P_n$ – условие обстановки, требуемые этим правилом для достижения цели, $PG_1 \& \dots \& PG_m$ – подцели, которые нужно достичь для достижения конечной цели $PG_0$ и $A_1 \& \dots \& A_k$ – действия, которые наряду с достижением подцелей, требуется выполнить, для достижения конечной цели $PG_0$. Когда некоторый интеллектуальный объект $\mathcal{C} := (C, Exp_C)$ «возводится в степень» имеющегося опыта с учетом обстановки, то он обогащается такими правилами, но без учеты действий, предполагая, что они будут осуществлены в будущем. Поэтому знания, которыми обогащается интеллектуальный объект о способе достижения цели имеют вид $P_1 \& \dots \& P_n \& PG_1 \& \dots \& PG_m \Rightarrow PG_0$, не содержащий действий. Это правило должно входить в вероятностное формальное понятие данного интеллектуального объекта.

На стадии принятия решений выбирается только один из способов достижения цели, формирующий план действий. Интеллектуальные объекты $\mathcal{C} := (C, Exp_C)$ «возведенные в степень» имеющегося опыта и включающие определенный способ достижения цели дают «тяжелые» контексты, соответствующие разным «образам потребного будущего». Среди этих «тяжелых» контекстов выбирается самый «тяжелый» $\mathcal{O} := (O, Exp_O)$ с наиболее желаемым «образом потребного будущего». Он и есть результирующий контекст первого этапа работы функциональной системы.

**Акцептор результатов действия**. Выбранный план действий, соответствующий выбранному «образу потребного будущего» включает в себя также последовательность и иерархию результатов, которые должны быть получены для достижения цели. Критерии достижения этих результатов, как совокупность определенных стимулов, которые должны быть получены при их достижении, формируют акцептор результатов действия. Это определенные

«интеллектуальные объекты» со своей стимуляцией, функцией ожидания и различительной способностью, фиксирующие достижение этих результатов.

**Второй этап** выполнения плана действий вместе с контролем достижения промежуточных и конечного результатов акцептором результатов действий выполняется в точном соответствии с полученным контекстом. В случае какого-либо отклонения от плана, включается ориентировочно исследовательская реакция, которая пересматривает план действий и возможно «образ потребного будущего» в результате чего сформированный контекст пересматривается.

Если в соответствии с выбранным планом действий цель достигается и потребность удовлетворяется, то этот план действий подкрепляется и заносится в память. В серии компьютерных экспериментов [27-33] была подтверждена работоспособность данной схемы.

## 8    Первый принцип ВРР – принцип генерации сложности.

Принцип генерации сложности в [1, стр 217] формулируется так: «Мозг работает с весьма ограниченным объемом информации от окружающей его реальности, поступающим на его сенсоры … По мере использования этой, изначально скудной информации мозг, на всех уровнях своей организации многократно увеличивает ее объем, соотнося полученные вводные с уже существующими в нем данными … Принцип генерации сложности позволяет мозгу, получив самый незначительный внешний сигнал, воспроизвести в сознании человека знание (интеллектуальный объект) несопоставимо большей мощности, обогатив модель этого объекта информацией, которая актуальна для мозга в рамках его задач (его целей)».

Такую генерацию сложности выполняет введенная ранее в [1, стр. 214] интеллектуальная функция, которая «в рассматриваемом нами контексте выступает единственным инструментом мышления, используя которую мы создаем новые отношения между интеллектуальными объектами».

Формализация «интеллектуального объекта» и «интеллектуальной функции» вероятностными формальными понятиями дает следующие модели генерации сложности:

1.  Если рассматривать признаки цифр рис. 1, как признаки, воспринимаемые первичной зрительной корой, а множество А, как множество воспринимаемых цифр, то множество вероятностных формальных понятий, которые были обнаружены для этих цифр порождает множество интеллектуальных объектов $\mathcal{A}_0 := (0, \mathrm{Exp}_0)$ , $\mathcal{A}_1 := (1, \mathrm{Exp}_1)$ ,…, $\mathcal{A}_9 := (9, \mathrm{Exp}_9)$ – инвариантов этих цифр. Причинные связи между признаки цифр на множестве А будут найдены автоматически, поскольку мозг всегда и везде обнаруживает причинные связи. Найденные инварианты – это пример сгенерированной сложности на базе простейших свойств.

2.  Формирование контекстов как вероятностных формальных понятий, что порождает, например, типологию пользователей социальной сети (см.

пример выше). Контексты могут быть разные, например, вербальный контекст в виде законченного отрывка текста, смысл которого уточняет значения входящих в него слов или ситуативный контекст, включающий обстановку, время, место и т.д., помогающий более точно интерпретировать значения высказываний об обстановке.

3. Формирование ситуативного контекста некоторой функциональной системой с целью формирования плана действий по удовлетворению некоторой потребности.

В общем случае, когда решается некоторая задача или достигается определенная цель генерация сложности соответствующей интеллектуальной функции будет состоять в генерации контекста по заданным начальным условиям A путем «возведения их в степень» тех знаний, которые имеют к ним прямое отношение. Формально это представляет собой генерацию некоторого вероятностного формального понятия по условиям A оператором $\Upsilon(A)^{\infty}$ с использованием всех относящихся к задаче или цели знаний, представленных совокупностью МСВПС правил.

# 9 Второй принцип ВРР – принцип отношения

В психологии этот принцип изначально получил название – принцип гештальта. Мозг, как мы знаем, реагирует не на конкретный стимул, а на то, каким становится этот стимул при соотнесении его с той информацией, которая в мозге уже содержится [1]. «Оценка возникающей в мозге информации … осуществляется исключительно через акт соотнесения одной информации с другой, а сам мозг реагирует не на объект реальности как таковой, а на то, как он соотносится с другой информацией, находящейся в мозге» [1].

В качестве основного объяснительного принципа гештальтпсихология выдвигает принцип целостности. «Целостность восприятия – свойство восприятия, состоящее в том, что всякий объект, а тем более пространственная предметная ситуация воспринимаются как *устойчивое системное целое*, даже если его некоторые части в данный момент нельзя наблюдать (например, тыльная часть вещи): актуально не воспринимаемые признаки всё же оказываются интегрированными в целостный образ этого объекта» (Википедия). Целостность восприятия, которая формируется в процессе восприятия «естественного» понятия или прототипа класса, а также контекста некоторой задачи, формально представляет в вероятностном формальном понятии взаимным предсказанием свойств понятия или элементов контекста. Поэтому вероятностное формальное понятие образует то самое «устойчивое системное целое», которое характеризует целостность.

Поэтому формально оператор $\Upsilon(A)^{\infty}$ и есть то самое «устойчивое системное целое», в котором, в котором воспринимаются не отдельные элементы $A$, а их неразрывная взаимосвязь с остальными элементами неподвижной точки $\Upsilon(A)^{\infty}$.

## 10 Третий принцип ВРР – принцип аппроксимации до сущности

Принцип аппроксимации в [1] описывается следующим образом: «… в реальности не существует абсолютно идентичных объектов, поэтому мозг осуществляет аппроксимацию, то есть игнорирует отличия, если ему удается по специфическим признакам присвоить объекту ту или иную «сущность». При этом, под «сущностью» понимается функционал объекта – то, какое значение он имеет для мозга (какую роль он выполняет) в рамках решаемых им задач (его целей). Наглядным примером в этом случае является использование какого-либо объекта в качестве другого, путем наделения первого функционалом второго под актуализированную потребность: когда человек устал и хочет отдохнуть – в лесу пень может служить стулом, так как на нем можно сидеть».

Формирование «сущностей» происходит в контексте решаемых задач или функциональных систем. Всякий контекст уточняет и взаимно соотносит элементы контекста. Это приводит к формированию «сущностей», связанных с контекстом. Например, нож в разных контекстах: приготовления пищи, боевой ситуации, офисной работы и походных условиях должен обладать разными свойствами, вытекающими из контекста: для кухонного ножа важна взаимосвязь ширины, веса и острия лезвия, для боевого ножа – соотношение острия, длины, веса и ширины, для канцелярского ножа – малость веса, длина и безопасность, для перочинного ножа – относительная малость размеров. Поэтому формируются «сущности» «кухонный нож», «боевой нож», «канцелярский нож», «перочинный нож», которые автоматически в соответствующих ситуациях порождают различные вероятностные формальные понятия, поскольку свойства и закономерности их взаимосвязи различны.

Контекст решаемой задачи, цели или потребности будет автоматически заставлять выбирать наиболее подходящие для этого объекты с соответствующим «функционалом». Этот функционал, имеющий определенное значение для мозга в рамках решаемых им задач и целей, определенным образом отразится на совокупности свойств объекта, которые, взаимно предполагая друг друга, автоматически сформируют соответствующее вероятностное формальное понятие, соответствующее его функциональной «сущности».

Поэтому «сущность» - это вероятностное формальное понятие $\Upsilon(A)^{\infty}$, порожденное такими элементами $A$ – свойствами используемых объектов, которые будут выбираться в соответствии с контекстом решаемой задачи или достигаемой цели.

## 11 Четвертый принцип ВРР – принцип локальности-распределенности (принцип симультанности)

Принцип локальности-распределенности [1]: «Вся информация, поступающая в мозг, может в нем многократно дублироваться, и ее копии обрабатываются параллельно разными структурами самостоятельно, и лишь затем эта информация интегрируется в целостный образ. Иными словами, мозг обрабатывает одну и ту же информацию разными способами (в разных отделах), чтобы получить несколько результатов и объединить их в рамках одного, целостного интеллектуального объекта, в соответствии с определенной им сущностью».

Мозг обрабатывает информацию о некотором объекте параллельно сразу в нескольких модальностях – зрительной, слуховой, тактильной и т.д. В каждой из этих модальностей образуется иерархия простейших «естественных» классов и понятий, например, в зрительной коре на основании воспринятых палочек могут формироваться образы цифр, как в приведенном выше примере, а также «вторичные» признаки – линии, углы, окружности и т.д., в слуховой коре – фонемы, слова, текст и т.д. Согласование модальностей образа осуществляется уже на верхнем уровне через восприятие целостности объекта, которая интегрирует и связывает восприятие частей в «устойчивое системное целое», что и осуществляют вероятностные формальные понятия целостных объектов.

Работа интеллектуальной функции по принципу локальности-распределённости (симультанности) состоит не только в том, чтобы интегрировать модальности некоторого образа и воссоздать интеллектуальные объекты, связанные с элементами воспринимаемых данных А, но и соединять эти интеллектуальные объекты со всеми другими интеллектуальными объектами, имеющимися в психике и имеющими отношение к данной ситуации, например, к некоторой потребности или задаче (цели). Таким образом, интеллектуальный объект $\mathcal{A} := (A, Exp_A)$ «как бы возводится в степень» тех знаний (интеллектуальных объектов), которыми мы обладаем и в результате работы интеллектуальной функции параллельно создаются «конусы» – множества $\{\mathcal{C} := (C, Exp_C)\}$ интеллектуальных объектов, как и случае функциональных систем, обогащающих исходный интеллектуальный объект $\mathcal{A} := (A, Exp_A)$ до некоторых целостных контекстов, определяющих возможные смыслы воспринимаемой ситуации А.

Поэтому формально этот принцип также представляется оператором $\Upsilon(A)^\infty$, генерирующим вероятностные формальные понятия целостных контекстов воспринимаемой ситуации А.

## 12    Пятый принцип ВРР – принцип тяжести

Принцип тяжести [1]: «Количество нейронных связей, включенных в создание модели объекта, количество отношений между элементами континуума интеллектуальных объектов, объем привносимой в объект информации (атрибуты сущности), количество способов расчета информации об объекте и объединение разноканальной (модальности) информации о нем в единое целое, соотнесенные с актуальностью задачи (цели) системы, определяют «тяжесть» интеллектуального объекта. «Тяжесть» интеллектуального объекта предопределяет решение системы. Так, например, если человек голоден – он будет искать пищу, которая утолит голод, однако если ему начнет угрожать непосредственная опасность (например, от хищника), то начнет главенствовать оборонительная стратегия, и он перестанет искать еду и начнет спасаться, так как без еды он проживет еще какое-то время, а если его настигнет хищник – он умрет сразу. То есть, приоритет отдается наиболее актуальной и выраженной в каждой конкретной ситуации стратегии».

Ещё в 1911 году А.А. Ухтомским был выдвинут принцип доминанты [34]. Он сохранился и в теории функциональных систем [3,4], как принцип доминирующей функциональной системы, которая и создает наиболее «тяжелый» контекст.

В общем случае, когда речь идет о решении некоторой задачи или достижении определенной цели, возможные решения по принципу локальности-распределенности (симультанности) получаются разными путями обогащения исходного интеллектуального объекта «постановка задачи» (цели) и образуют соответствующие «конусы» и порождаемые ими контексты. Выбор из них самого «тяжелого» определяется выбором наиболее желаемого «тяжелого» решения.

Поэтому формально принцип тяжести состоит в выборе наиболее желаемого «тяжелого» интеллектуального объекта порожденного одним из контекстов, которые генерируются оператором $\Upsilon^{*}(M \cup C)$ в зависимости от исходной постановки задачи/цели $\mathcal{M} := (M, Exp_M)$ и имеющегося опыта $C$ решения подобной задачи/цели.

## 13    Системная взаимосвязь принципов

Когда мы берем в руки яблоко (см. рис. 4), то получаем первичную информацию $A$ о нем – оно твердое, имеет средний вес, форма круглая, поверхность гладкая, размер средний. Сначала начинает работу первый принцип – генерации сложности: «Мозг работает с весьма ограниченным объемом

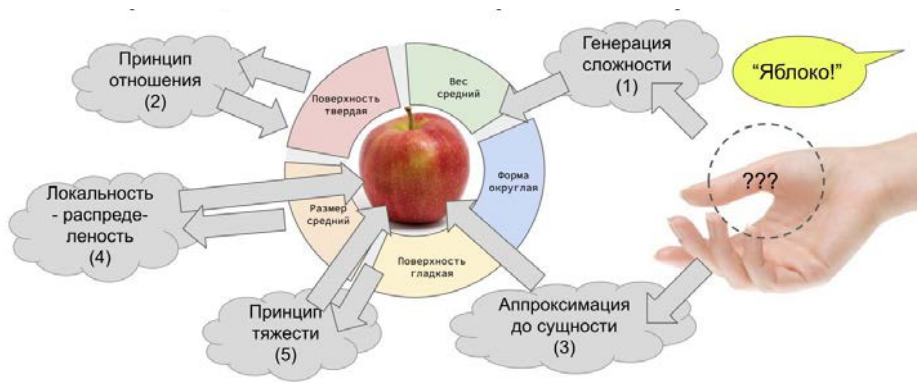

**Рис. 4.** Системная взаимосвязь принципов.

информации от окружающей его реальности, поступающим на его сенсоры … По мере использования этой, изначально скудной информации мозг, на всех уровнях своей организации многократно увеличивает ее объем, соотнося полученные вводные с уже существующими в нем данными …» [1]. Поэтому мы сразу понимаем, что это не апельсин, поскольку апельсин не имеет гладкой поверхности и это не мячик, т.к. мячики не твердые и это не бильярдный шар, поскольку они тяжелые и это не теннисный мячик, поскольку его поверхность не шершавая и.т.д.

Далее работает принцип отношений: «Оценка возникающей в мозге информации … осуществляется исключительно через акт соотнесения одной информации с другой, а сам мозг реагирует не на объект реальности как таковой, а на то, как он соотносится с другой информацией, находящейся в мозге» [1], «всякий объект… воспринимаются как устойчивое системное целое, даже если его некоторые части в данный момент нельзя наблюдать» (Википедия). Воспринятая информация должна соотносится сама с собой и образовывать некоторое системное целое – поэтому из имеющихся в памяти образов извлекается именно образ яблока и этот образ также дополнительными (перцептивными) действиями проверяется на целостность, например, на наличие хвостика и диаметрально расположенных ямок.

Далее по принципу генерации сложности, который работает всегда, воспринятая информация ещё более обогащается имеющимися знаниями и возникает понимание «сущности» воспринятого яблока по третьему принципу аппроксимации до сущности: под ««сущностью» понимается функциональ объекта – то, какое значение он имеет для мозга (какую роль он выполняет) в рамках решаемых им задач» [1]. Это яблоко «съедобное», если мы собираемся его есть, это яблоко «красивое», если мы им просто любуемся или собираем натюрморт, это яблоко «сортовое», если нам нужны его косточки для дальнейшего разведения и т.д.

Далее, все также по принципу генерации сложности воспринятая информация еще более обогащается «возводится в степень» интеллектуальной функцией параллельно по разным каналам обработки информации вплоть до целостного её восприятия в соответствии с четвертым принципом локальности-распределенности (симультанности): «Вся информация, поступающая в мозг, может в нем многократно дублироваться, и ее копии обрабатываются параллельно разными структурами самостоятельно, и лишь затем эта информация интегрируется в целостный образ» [1]. Этот целостный образ формирует контекст восприятия – в каком контексте мы воспринимаем яблоко – в контексте поесть, в контексте полюбоваться, в контексте семеноводства и т.д.

Пятый принцип – принцип тяжести определяет наш выбор наиболее тяжелого контекста из воспринятых, который в данный момент наиболее желаем для нас или больше всего нас интересует и который определит наши дальнейшие размышления или действия: «объединение разноканальной информации в единое целое, соотнесенные с актуальностью задачи (цели) системы, определяют «тяжесть» интеллектуального объекта. «Тяжесть» интеллектуального объекта предопределяет решение системы» [1].

С формальной точки зрения непрерывная работа интеллектуальной функции в соответствии с принципом генерации сложности по обогащению воспринятой информации описывается оператором $\Upsilon(A)$, интеллектуальные объекты и контексты, получаемые по второму, третьему и четвертому принципам, описываются вероятностными формальными понятиями, т.е. оператором $\Upsilon^\infty(A)$. Устойчивое системное целое, характеризующее целостность этих интеллектуальных объектов следует из строения вероятностных формальных понятий, выявляющих системный закон (в виде согласованной системы причинных связей) строения интеллектуальных объектов. Следует отметить, что интеллектуальные объекты, получаемые вероятностными формальными понятиями, описываются не только как набор (синдром) признаков, но и как устойчивое системное целое «инвариант», характеризующийся системной взаимосвязью причинных связей, взаимно предсказывающих признаки интеллектуального объекта.

В основе формализации BPP в рамках теории категорий также лежит только два понятия – «интеллектуальный объект» и «интеллектуальная функция», функционирование которых разворачивается в принципах. В нашей формализации интеллектуальному объекту соответствует оператор $\Upsilon^\infty(A)$, а интеллектуальной функции – оператор $\Upsilon(A)$.

В работе [34] приведена англоязычная версия описания базовых моделей и принципов программирования мозга.

## 14    Практические соображения и применимость в прикладных задачах

Алгоритмы построения вероятностных формальных понятий и соответственно прототипов классов, «естественных» понятий и контекстов практически подтверждены [12,16-20,24]. Модель функциональных систем также разработана и показала свою эффективность [26-32]. Интеграция этих алгоритмов и моделей может быть осуществлена путем использования в контекстах функциональных систем правил без действий (см. раздел 6) в предположении, что необходимые действия будут выполнены и проконтролированы в соответствии с включенными в контекст функциональными системами. Интегрированный алгоритм достаточно точно моделирует основные когнитивные функции человека и животных, упомянутые в первых частях статьи, поэтому область применимости может быть широка. Для этого необходимо провести масштабирование данного подхода.

## 15    Выводы

Данный подход может быть обобщен до *задачного подхода к общему искусственному интеллекту*, как это и планировалось в [1] путем обобщения функциональных систем до систем решения задач [35-39]. Этот подход вполне справляется с задачей AGI, сформулированной в [1] как: «способность достигать целей в широком диапазоне сред с учетом ограничений».

Поэтому Brain Principles Programming, сформулированные в [1] как принципы программирования мозга, опираясь на исследования в когнитивных науках, могут быть реализованы как задачный подход к AGI, который одновременно способен решать достаточно широкий класс задач, а, с другой стороны, достаточно точно соответствует моделям когнитивных процессов.

## Литература